\documentclass{article}
\usepackage{spconf,amsmath,graphicx}
\usepackage{xcolor}
\usepackage{multirow}
\usepackage{enumitem}
\usepackage{amssymb}
\usepackage{tikz}       
\usepackage{fancyhdr}      
\pdfoutput=1

\usetikzlibrary{calc,arrows,positioning,shapes,shadows,spy,plotmarks,matrix,fit,backgrounds}

\definecolor{maize}{rgb}{0.98, 0.93, 0.37}

\title{MDA-Net: Multi-Dimensional Attention-Based \\ Neural Network for 3D Image Segmentation}
\name{Rutu Gandhi, Yi Hong}
\address{Institute of AI, University of Georgia, Athens}
\twoauthors
 {Rutu Gandhi}
	{Institute for Artificial Intelligence\\
	University of Georgia, Athens, GA, USA}
 {Yi Hong\sthanks{yi.hong@sjtu.edu.cn}}
	{Dept. of Computer Science and Engineering\\
	Shanghai Jiao Tong University, Shanghai, China}

\begin{document}
%\ninept
%
\maketitle
%
% Patch-based segmentation using expert priors: Application to hippocampus and ventricle segmentation
\begin{abstract}
Segmenting an entire 3D image often has high computational complexity and requires large memory consumption; by contrast, performing volumetric segmentation in a slice-by-slice manner is efficient but does not fully leverage the 3D data. To address this challenge, we propose a multi-dimensional attention network (MDA-Net) to efficiently integrate slice-wise, spatial, and channel-wise attention into a U-Net based network, which results in high segmentation accuracy with a low computational cost. We evaluate our model on the MICCAI iSeg and IBSR datasets, and the experimental results demonstrate consistent improvements over existing methods. 

%The abstract should appear at the top of the left-hand column of text, about
%0.5 inch (12 mm) below the title area and no more than 3.125 inches (80 mm) in
%length.  Leave a 0.5 inch (12 mm) space between the end of the abstract and the
%beginning of the main text.  The abstract should contain about 100 to 150
%words, and should be identical to the abstract text submitted electronically
%along with the paper cover sheet.  All manuscripts must be in English, printed
%in black ink.
\end{abstract}
\section{Introduction}
\label{sec:intro}

\label{sec:intro}
Image segmentation is a fundamental task in image understanding, which distinguishes regions of interest from image background for further analysis. Recently, deep segmentation networks, e.g., fully convolutional networks (FCN)~\cite{long2015fully}, U-Net~\cite{ronneberger2015u}, tackle the 2D image segmentation problem and outperform conventional approaches. However, segmenting 3D image volume like brain MRI scans is still challenging, especially in medical image analysis. 
%Apart from 2D images, segmentation techniques are also desired for other commonly-used 2D+ data types, e.g., 3D image volumes like brain MRI scans, videos, and longitudinal images. 
Models extended from FCNs and U-Nets have been proposed to handle the 3D image segmentation, e.g., V-Net~\cite{milletari2016v}. Due to the high-dimensional nature of image data, most existing models have a high demand for computational resources, especially the GPU memory, and often have a large number of parameters to estimate. 

\begin{figure*}[h]
\centering
\includegraphics[width=0.74\textwidth]{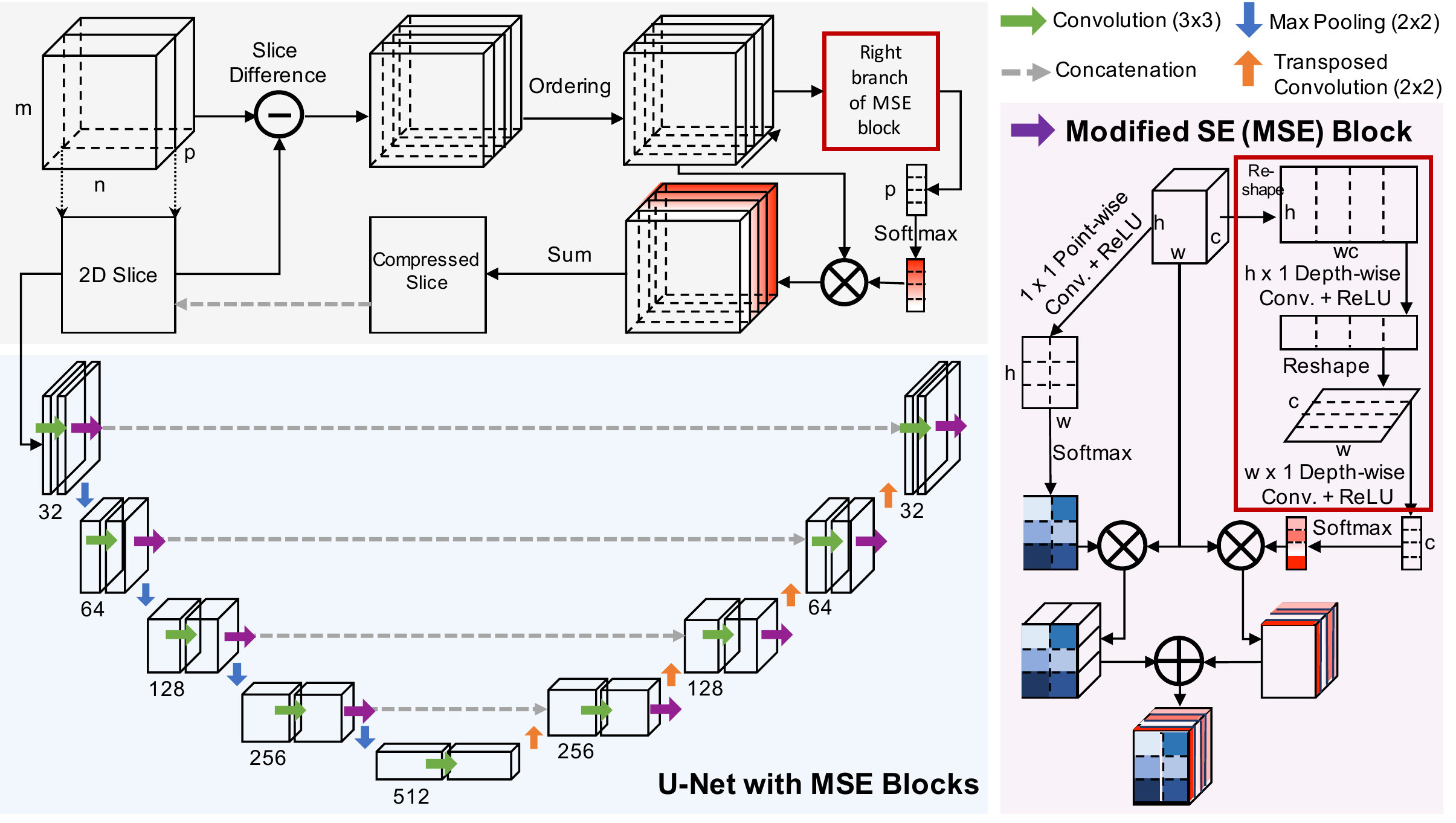}
\caption{Overview of our proposed multi-dimensional attention network (MDA-Net) for 3D image segmentation. It consists of a third-dimensional compression and an attention-augmented U-Net with modified squeeze and excitation (MSE) blocks.} \label{fig:overview}
\end{figure*}

To integrate the information across slices of an image volume, we propose a compression technique based on the squeeze and excitation (SE) technique~\cite{hu2018squeeze} to concisely abstract multiple neighboring slices of a volume into a single slice. Apart from the augmented input for the 2D backbone network, we also augment the network with spatial and channel-wise attention by upgrading the concurrent SE block~\cite{roy2018concurrent}. The resulting network benefits from both the low computation cost of the 2D network and enriched information from image volumes and learned features with attention. Figure~\ref{fig:overview} depicts the overall architecture of our multi-dimensional attention network (MDA-Net) for segmenting 3D images. The MDA-Net is an end-to-end solution and can automatically learn how to compress volumetric image information and extract useful features in an attention scheme. 

% The ordered volume is given as input to the right branch of MSE only once outside the U-Net
Our MDA-Net mimics the process of the manual segmentation for a 3D image. When handling an image volume, we often select one main view to sequentially segment 2D slices and check the third dimension across slices occasionally to obtain additional information. To integrate the information among image slices, we condense the ordered slice difference computed with respect to the current main slice. This compression step allows collecting extra information, i.e., image residuals, to assist the segmentation. Another benefit of using our model is the increased data samples. If we have $p$ slices in one view, we convert one 3D image into at most $p$ samples. Each sample contains a 2D slice and another slice compressed from the original image. This compression is achieved by using slice-wise attention. We also have the spatial and channel-wise attention used in the segmentation network, resulting in our multi-dimensional attention network. The attention is automatically estimated via modified squeeze and excitation (MSE) blocks, which improve segmentation performance over the original SE block and concurrent scSE block.

The main contributions of this paper are as follows:
\begin{itemize}[noitemsep, nolistsep]
    \item Modified SE (MSE) Block: We upgrade the channel attention mechanism in the concurrent scSE and replace the sigmoid function with softmax to ensure the weights for measuring the attention are normalized.
    \item Slice-wise condensing module: We propose a compression module that uses slice-wise attention to extract residual information in the third dimension.
    \item Multi-Dimension Attention Network (MDA-Net): We propose an efficient 3D image segmentation network, which fully leverages the 3D data with the balanced computational cost.  
\end{itemize}

We evaluate our MDA-Net on the MICCAI iSeg dataset~\cite{wang2019benchmark} and the IBSR dataset~\cite{rohlfing2011image} through segmenting 3D brain scans. The segmentation results on both datasets show the improvement over previous methods with five-fold cross-validation.

%\subsection{Related Work}
%\label{related_work}

\noindent
\textbf{Related Work.}
The 3D variants of the U-Net~\cite{milletari2016v,cciccek20163d} were proposed to handle volumetric images. Compared to a 2D U-Net working slice by slice, its 3D version fully uses the data in all dimensions. However, 3D models face two main challenges in segmenting the entire high-dimensional image volume. We often have limited computational resources, especially limited GPU memory, to handle the whole image volume. Compared to a greatly increased number of model parameters when switching from a 2D network to 3D, we have a reduced number of data samples since a 2D slice sequence becomes one sample in a 3D U-Net. Existing approaches to address these challenges include downsampling the 3D images to fit in memory~\cite{cciccek20163d}, assembling multiple 2D networks for accepting different image views~\cite{chlebus2018automatic}, working on 3D image patches~\cite{liu2019point}, modifying the existing segmentation architectures~\cite{lucas2018multi, brugger2019partially} or combining 2D slices with 3D patches~\cite{dey2019hybrid}.

Unlike previous approaches, we segment an image volume in the slice-to-slice fashion while integrating the residuals across slices. A work related to ours is the volumetric attention Mask-RCNN~\cite{wang2019volumetric}, which considers three adjacent slices when calculating attention. Our attention model is built based on the SE block, which is not limited to three slices and provides the first-order statistic across slices.

\section {MDA-Net}
To provide an economical solution for 3D image segmentation, we adhere to treat the 3D data as a sequence of 2D image slices, which allows us to work in a lower-dimensional space with relatively low demand in computational resources. For a specific 2D slice, we augment it with the first-order information in the third dimension, which is achieved by condensing ordered image differences in its neighborhood into one slice. Our model in Fig.~\ref{fig:overview}  has an image dimension reduction component using an ordering-based image difference compression. The resulting slice is concatenated with the associated 2D slice as inputs for an attention-augmented 2D U-Net. We stack the 2D segmentation masks back to form the segmentation mask for an image volume.  

\subsection{Slice-wise Compression}
\label{sec:compression}
Assume we segment the $i$-th image slice, e.g., $I_{m\times n}^{(i)}$, of an image volume $I_{m\times n \times p}$, $i = 1, \cdots, p$. Besides the 2D slice, i.e., $I_{m\times n}^{(i)}$, we prepare an additional slice $\bar{I}_{m\times n}^{(i)}$ that associates with $I_{m\times n}^{(i)}$ and contains information across the slices. The concatenation, $I_{m\times n}^{(i)} \cup \bar{I}_{m\times n}^{(i)}$, is the input of our attention-augmented U-Net described in Section~\ref{sec:u-net}. This new presentation allows us to take full use of the data in all three dimensions when segmenting one 2D slice of the 3D image volume. 

A typical way to compress the data from 3D to 2D is by performing a weighted average over the third dimension. Instead of directly compressing the original image volume, we choose to compress the difference images $I_{m\times n}^{(j+i)} - I_{m\times n}^{(i)}$, $j \in [-r, \cdots, -1, 1, \cdots, r]$. Here, $2r$ (e.g., $r=5$ chosen experimentally) images in the 2D slice neighborhood are selected. A weighted average of these difference images is the condensed slice $\bar{I}_{m\times n}^{(i)}$, and the weights are estimated using the right branch of our MSE block, as described in the next paragraph. As the center image $I_{m\times n}^{(j)}$ changes, the difference images will change accordingly, and their contributions in the estimated condensed image, which are measured by their weights, vary as well. However, once the network is trained, the weights for the difference images are fixed. To have the weights consistently associated with the difference images, we order them based on the sum of their absolute values. That is, the compressed slice $\bar{I}_{m\times n}^{(j)}$ is the weighted average of the ordered difference images, as shown in Fig.~\ref{fig:overview}. 

To estimate the weights for the ordered difference images, we follow the spatial squeeze and channel excitation idea~\cite{hu2018squeeze} but with some modifications on channel-wise SE block. Firstly, the original SE block squeezes the spatial domain using the global average pooling. That is, the SE block summarizes each channel using its average over pixels with equal weights. The foreground's pixels are often sparse in the difference images, and nearly-zero pixels dominate in the background. Therefore, we need the flexibility in spatial squeeze and adopt convolution filters to average over pixels with learned weights. A simple way to get a weight for each difference image (or a channel used later) is to use a filter of size $m \times n$. To reduce the number of parameters, we decompose this 2D filter into two 1D filters, i.e., one with size $m \times 1$ and the other with size $n \times 1$, and apply depth-wise convolutions. In this way, we can reduce the number of parameters from $mn$ to $m + n$. Since we expect to summarize the information in each channel independently, we need to reshape the input of each depth-wise convolution accordingly, as shown in Fig.~\ref{fig:overview}. As a result, we obtain a vector $z \in \mathbb{R}_{p}$ as the unnormalized weights for difference images. Also, instead of using Sigmoid in~\cite{hu2018squeeze}, which normalizes each weight independently into the range [0, 1], we choose the Softmax function, which counts the correlation among weights and enforces their sum to be 1. The normalized weights are used to rescale the ordered difference images, and then a weighted average results in the compressed slice $\bar{I}_{m\times n}^{(i)}$.

\subsection{2D U-Net with MSE Blocks}
\label{sec:u-net}
After compressing the residual information in the third dimension, we represent a 3D volume as a pair of 2D slices, i.e., the current 2D slice $I_{m\times n}^{(j)}$ and the slice-wise compression $\bar{I}_{m\times n}^{(j)}$. A 2D U-Net takes our 2D slice pairs as a two-channel input for segmenting the current 2D image. Using a similar approach in the slice-wise compression, which offers us the third dimension's attention, we add MSE blocks to a plain 2D UNet, which provides both spatial and channel-wise attention on its input and extracted feature maps. 

Like the concurrent SE block in~\cite{roy2018concurrent}, we add both channel squeeze and spatial excitation branch (sSE) and spatial squeeze and channel excitation branch (cSE), as shown in Fig.~\ref{fig:overview}. But, different from~\cite{roy2018concurrent}, we use the Softmax operator instead of Sigmoid to normalize the spatial or channel-wise weights in both branches. We consider the attention depends on the space and the channels, and the separated weights allow us to treat spatial pixels or channels differently. In particular, the sSE branch uses a pixel-wise convolution to summarize the input feature maps across channels, which is then normalized over the spatial domain using Softmax to ensure the weights' sum is 1. The cSE branch is the same as we discussed in Section~\ref{sec:compression}. In particular, given an input feature map $U_{H \times W \times C}$ ($H \times W$ are the spatial size of the feature map and $C$ is the number of channels), we apply depth-wise convolution with an $H \times 1$ filter followed by a $W \times 1$ filter. The resulting $C \times 1$ vector is normalized using Softmax to rescale the channels before taking their average. Reshaping the feature maps from 3D to 2D or from 1D to 2D is required before applying the depth-wise convolutions. The addition of these two branches gives us the output of the MSE block. Each MSE block is used after the convolution pair at each resolution of the U-Net. Worth to mention that, different from the concurrent SE block, we do not use fully-connected layers, resulting in a drop in the number of model parameters.  

%In this way, we completely avoid the module that uses two dense layers in the concurrent SE architecture. As seen in~\ref{tab:computational_cost}, this results in a significant drop in the number of trainable parameters. The network architecture details are shown in Fig.~\ref{fig:overview}.

%New arch: To further reduce the number of parameters, we first flatten the $U_{H \times W \times C}$ block into $U_{H \times WC}$ and then apply DepthwiseConv2D with filter size $H \times 1$. This reshaping step allows us to use one depthwise filter for every channel of the 3D block, thus reducing the number of parameters by a factor of 10. The output of this layer is a vector of size $U_{1 \times WC}$ for each sample. After this, we restack the vector into a $U_{W\times C}$ dimensional 2D array.https://www.overleaf.com/project/5d8e5978568f1d0001da3de8 Anothhttps://www.overleaf.com/project/5d8e5978568f1d0001da3de8er depthwise layer with a filter of size $1 \times W$ is then applied to the the 2D array which converts it into a vector of size $C$.

\begin{table*}[t]
\begin{center}
\small
 \begin{tabular}{l|lll|lll|lll} 
 \hline
 & \multicolumn{3}{c|}{Sagittal (\%)} & \multicolumn{3}{c|}{Axial (\%)} & \multicolumn{3}{c}{Coronal (\%)}\\
 \cline{2-10}
 & \footnotesize{CSF} & \footnotesize{GM} & \footnotesize{WM} & \footnotesize{CSF} & \footnotesize{GM} & \footnotesize{WM} & \footnotesize{CSF} & \footnotesize{GM} & \footnotesize{WM} \\ 
 \hline
 Plain UNet~\cite{ronneberger2015u} & \footnotesize{91.41}\scriptsize{$\pm$1.12} & \footnotesize{81.54}\scriptsize{$\pm$1.37} & \footnotesize{75.13}\scriptsize{$\pm$1.19} & \footnotesize{95.22}\scriptsize{$\pm$1.01} & \footnotesize{92.33}\scriptsize{$\pm$1.18} & \footnotesize{89.51}\scriptsize{$\pm$2.16} & \footnotesize{92.48}\scriptsize{$\pm$2.34} & \footnotesize{84.36}\scriptsize{$\pm$1.16} & \footnotesize{80.24}\scriptsize{$\pm$2.35} \\ 
 cscSE-UNet~\cite{roy2018concurrent} & \footnotesize{92.71}\scriptsize{$\pm$1.74} & \footnotesize{85.56}\scriptsize{$\pm$1.14} & \footnotesize{79.14}\scriptsize{$\pm$1.11} & \footnotesize{96.47}\scriptsize{$\pm$1.01} & \footnotesize{92.38}\scriptsize{$\pm$1.18} & \footnotesize{89.83}\scriptsize{$\pm$2.13} & \footnotesize{92.11}\scriptsize{$\pm$1.45} & \footnotesize{87.12}\scriptsize{$\pm$1.09} & \footnotesize{82.15}\scriptsize{$\pm$1.42} \\
 MSE-UNet \footnotesize{(Ours)} & \footnotesize{93.11}\scriptsize{$\pm$1.34} & \footnotesize{{ 87.21}}\scriptsize{{$\pm$1.54}} & \footnotesize{81.13}\scriptsize{$\pm$2.34} & \footnotesize{96.27}\scriptsize{$\pm$1.01} & \footnotesize{92.62}\scriptsize{$\pm$1.18} & \footnotesize{89.89}\scriptsize{$\pm$2.13} & \footnotesize{93.91}\scriptsize{$\pm$2.34} & \footnotesize{88.25}\scriptsize{$\pm$2.16} & \footnotesize{84.23}\scriptsize{$\pm$2.65} \\
MDA-Net \footnotesize{(Ours)}  & \footnotesize{{\bf 93.46}}\scriptsize{{\bf $\pm$1.05}} & \footnotesize{{\bf 87.23}}\scriptsize{{\bf $\pm$2.64}} & \footnotesize{{\bf 82.79}}\scriptsize{{\bf $\pm$2.55}} & \footnotesize{{\bf 96.81}}\scriptsize{{\bf $\pm$1.15}} & \footnotesize{{\bf 94.32}}\scriptsize{{\bf $\pm$2.13}} & \footnotesize{{\bf 90.01}}\scriptsize{{\bf $\pm$1.24}} & \footnotesize{{\bf 94.28}}\scriptsize{{\bf $\pm$2.01}} & \footnotesize{{\bf 88.44}}\scriptsize{{\bf $\pm$1.45}} & \footnotesize{{\bf 85.23}}\scriptsize{{\bf $\pm$1.67}} \\
 \hline
\end{tabular}
\caption{Segmentation comparison (measured in Dice score) among different approaches applied on the MICCAI iSeg dataset. }
\label{tab:5_fold_average_iseg}
\end{center}
\end{table*}

\begin{table*}[h]
\begin{center}
\small
 \begin{tabular}{l|lll|lll|lll} 
 \hline
 & \multicolumn{3}{c|}{Sagittal (\%)} & \multicolumn{3}{c|}{Axial (\%)} & \multicolumn{3}{c}{Coronal (\%)}\\
 \cline{2-10}
 & \footnotesize{CSF} & \footnotesize{GM} & \footnotesize{WM} & \footnotesize{CSF} & \footnotesize{GM} & \footnotesize{WM} & \footnotesize{CSF} & \footnotesize{GM} & \footnotesize{WM} \\ 
 \hline
 Plain UNet~\cite{ronneberger2015u} & \footnotesize{90.81}\scriptsize{$\pm$1.85} & \footnotesize{95.91}\scriptsize{$\pm$2.24} & \footnotesize{92.92}\scriptsize{$\pm$1.55} & \footnotesize{91.24}\scriptsize{$\pm$1.15} & \footnotesize{90.33}\scriptsize{$\pm$2.16} & \footnotesize{88.38}\scriptsize{$\pm$1.84} & \footnotesize{91.45}\scriptsize{$\pm$2.08} & \footnotesize{87.30}\scriptsize{$\pm$1.46} & \footnotesize{89.22}\scriptsize{$\pm$2.67} \\ 
cscSE-UNet~\cite{roy2018concurrent} & \footnotesize{92.41}\scriptsize{$\pm$1.15} & \footnotesize{96.06}\scriptsize{$\pm$2.14} & \footnotesize{93.61}\scriptsize{$\pm$2.55} & \footnotesize{91.42}\scriptsize{$\pm$2.15} & \footnotesize{91.17}\scriptsize{$\pm$2.10} & \footnotesize{88.81}\scriptsize{$\pm$1.04} & \footnotesize{92.15}\scriptsize{$\pm$1.01} & \footnotesize{88.12}\scriptsize{$\pm$1.95} & \footnotesize{89.82}\scriptsize{$\pm$1.27} \\
MSE-UNet \footnotesize{(Ours)} & \footnotesize{94.01}\scriptsize{$\pm$1.95} & \footnotesize{96.96}\scriptsize{$\pm$2.14} & \footnotesize{94.79}\scriptsize{$\pm$3.55} & \footnotesize{92.23}\scriptsize{$\pm$1.25} & \footnotesize{92.65}\scriptsize{$\pm$2.14} & \footnotesize{89.82}\scriptsize{$\pm$1.64} & \footnotesize{93.93}\scriptsize{$\pm$2.01} & \footnotesize{89.21}\scriptsize{$\pm$1.15} & \footnotesize{90.58}\scriptsize{$\pm$1.60} \\
MDA-Net \footnotesize{(Ours)}  & \footnotesize{{\bf 95.12}}\scriptsize{{\bf $\pm$1.00}} & \footnotesize{{\bf 97.27}}\scriptsize{{\bf $\pm$2.14}} & \footnotesize{{\bf 95.04}}\scriptsize{{\bf $\pm$1.15}} & \footnotesize{{\bf 94.89}}\scriptsize{{\bf $\pm$1.05}} & \footnotesize{{\bf 95.31}}\scriptsize{{\bf $\pm$2.73}} & \footnotesize{{\bf 92.01}}\scriptsize{{\bf $\pm$2.24}} & \footnotesize{{\bf 94.52}}\scriptsize{{\bf $\pm$2.31}} & \footnotesize{{\bf 91.45}}\scriptsize{{\bf $\pm$1.35}} & \footnotesize{{\bf 92.22}}\scriptsize{{\bf $\pm$1.67}} \\
 \hline
\end{tabular}
\caption{Segmentation comparison (measured in Dice score) among different approaches applied on the IBSR dataset.}
\label{tab:5_fold_average_IBSR}
\end{center}
\end{table*}

\begin{figure}[t]
\centering
\begin{tikzpicture}[thick, spy using outlines={circle,lens={scale=2}, width=1.5cm, height=1.5cm, connect spies}]
	\node (reg_id1) {\includegraphics[width=0.99\columnwidth]{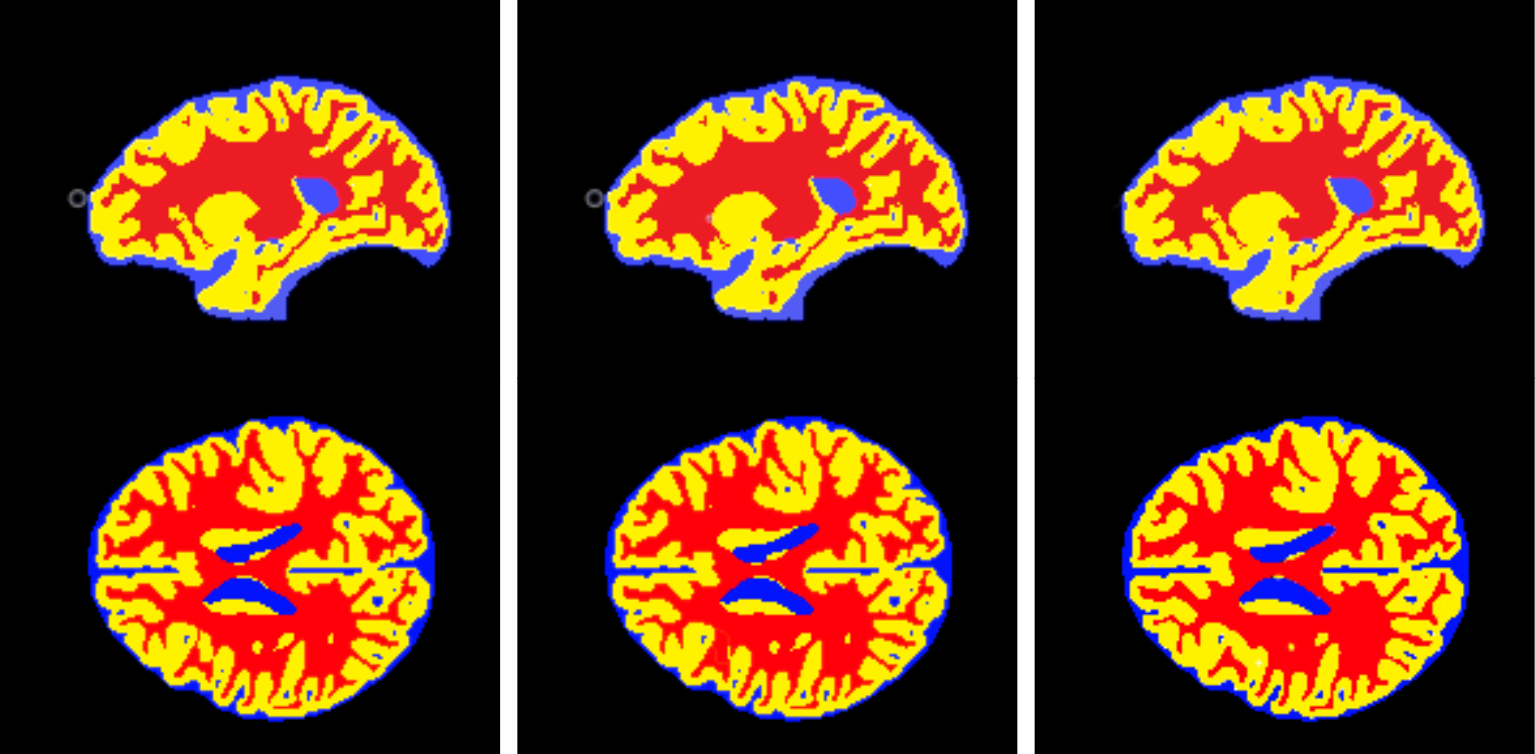}};
 	\spy [white] on (-3.2, 0.9) in node (lspy) [left,fill=black!5] at (-1.5,2.5);
 	\spy [white] on (-0.33, 0.9) in node (lspy) [left,fill=black!5] at (1.37,2.5);
 	\spy [white] on (2.54, 0.9) in node (lspy) [left,fill=black!5] at (4.24,2.5);
 	\spy [white] on (-3.2, -1.6) in node (lspy) [left,fill=black!5] at (-2.65,-2.7);
 	\spy [white] on (-0.33, -1.6) in node (lspy) [left,fill=black!5] at (0.22,-2.7);
 	\spy [white] on (2.54, -1.6) in node (lspy) [left,fill=black!5] at (3.09,-2.7);
\end{tikzpicture}
\caption{Qualitative iSeg sample results (from left to right: Ground-Truth, cscSE-UNet, and our MDA-Net. {\bf \color{blue} Blue}: cerebrospinal fluid, {\bf \color{yellow} yellow}: gray matter, {\bf \color{red} red}: white matter. } 
\label{fig:visual_results}
\end{figure}

\section{Experiments}
\subsection{Datasets and Experimental Settings}

\noindent
\textbf{iSeg Dataset.}
We evaluate our model on the dataset provided by the MICCAI iSeg challenge~\cite{wang2019benchmark}, which aims to segment the brain MRI scans into the cerebrospinal fluid (CSF), white matter (WM), gray matter (GM), and the background. The dataset consists of T1-weighted brain MRI scans collected from 10 6-month infants. Each scan has an image dimension of $144 \times 256 \times 192$ with a spatial resolution of $1\times 1\times 1 mm^3$. 

\noindent
\textbf{IBSR Dataset.}
We use the IBSR dataset~\cite{rohlfing2011image} to further demonstrate the performance improvement of our model over other approaches. The MR images provided in this dataset are T1-weighted three-dimensional coronal brain scans preprocessed with a positional normalization and a skull-stripping step. It contains 18 subjects with an image dimension of $512 \times 128 \times 256$ and a spatial resolution of 1.5mm in each dimension. 

%scans with segmentation and skull-stripped in NIfTI data format.

%The segmentation files are the result of semi-automated segmentation techniques which require many hours of effort by a trained expert. 

\noindent
{\bf Ablation Study.} The backbone network of our method is the 2D U-Net~\cite{ronneberger2015u}, which is augmented by our MSE blocks. In the ablation study, we compare our MSE-UNet with both the plain UNet and the concurrent scSE-UNet (cscSE-UNet)~\cite{roy2018concurrent}. These three models take only a single 2D slice of a 3D image scan to perform the 3D segmentation task in a slice-by-slice manner. Our whole model (MDA-UNet) is also reported to demonstrate the performance gain by including the slice-wise compression component into the MSE-UNet. 

\vspace{0.03in}
\noindent
%{\bf Other Settings.} 
We examine three views for both datasets, i.e., sagittal, axial, and coronal, respectively. Take the iSeg dataset for instance, we have 1152 samples (including a 2D slice and a sequence of difference images) in the sagittal view, 2048 samples in the axial view, and 1536 samples in the coronal view.

%The Plain U-Net, Concurrent scSE-UNet, our Modified scSE-UNet and our MDA-net are compared by performing 5-fold cross validation on each of the three views: Sagittal, Axial and Coronal for each of the 4 above-mentioned models. Table ~\ref{tab:5_fold_average_iseg} reports the average over the 5 folds. For the iSeg dataset, the 10 subjects are randomly shuffled and split into training and test sets to perform five-fold cross validation. For the IBSR dataset, the cross-validation is performed in a similar way on the 18 subjects. Table ~\ref{tab:5_fold_average_IBSR} reports the results for IBSR.

% \noindent
% {\bf Challenge Test Set.} We select the Axial view based on the results of the ablation study and cross validation, train on all 10 image scans, and submit the prediction results for the test set to the challenge. {\color{red}***We do need this result because of we are working a public available dataset; otherwise, we need another dataset to demonstrate the advantages of our method ****}

We implement our model using Keras and the Tensorflow backend. We use Adam optimizer with a learning rate of 5e-5, the dropout of 0.3, and the L2 regularizer. The Dice score is used during training and evaluation. The maximum number of epochs is 300. All models were trained on one NVIDIA GeForce 1080 8GB GPU. We use five-fold cross-validation on the subjects for both datasets. 
% Given a 3D brain MRI scan the U-Net accepts the 2D slices as input, generates mask on them which are stacked back into a 3D mask. On the other hand, our proposed architecture applies the difference and order operators to 5 slices before and 5 slices after the current slice for each slice. This block is then passed to the model.

% On the other hand, the U-CLSTM gives 3D brain scans at once to the ConvLSTM and then reshapes them into the corresponding 2D versions to give to the remaining encoder block. A Conv-LSTM is added to the beginning of each encoder block.

\begin{table}[t]
\begin{center}
\footnotesize
\begin{tabular}{l|l|c|c} 
 \hline
 & \multirow{2}{*}{\small{\#Param.}} & \small{Training time} & \small{Inference time} \\
 & & (per epoch) & (per patient)\\
\hline
 Plain UNet & 7.775M & 30s & 1.565s \\ 
 cscSE-UNet & 7.958M & 59s & 1.575s\\
 MSE-UNet & 7,823M & 57s & 2.285s\\ MDA-Net & 7.825M & 58s & 2.795s\\ 
 \hline
\end{tabular}
\caption{Model comparison on the number of model parameters and the training and testing time for the iSeg dataset.}
\label{tab:computational_cost}
\end{center}
\end{table}

\subsection{Experimental Results}
%The Multi-Dimensional Attention U-Net is compared to the Plain U-Net, the Concurrent Spatial and Channel Squeeze U-Net as introduced by \cite{roy2018concurrent} and its modified version. 
Tables~\ref{tab:5_fold_average_iseg} and~\ref{tab:5_fold_average_IBSR} report the segmentation results for iSeg and IBSR datasets, respectively. The brain segmentation performance for both datasets is steadily improved from plain U-Net, to cscSE-UNet, to MSE-UNet, and then to MDA-Net for each image view. The improvement of MSE-UNet over cscSE-UNet demonstrates the effectiveness of using MSE-Blocks in U-Net, and the improvement of MDA-Net over MSE-UNet demonstrates the effectiveness of using the slice-wise compression. Also, replacing Sigmoid with Softmax improves the stability and convergence of the network training. Figure~\ref{fig:visual_results} shows a qualitative comparison for the iSeg dataset. 

Table~\ref{tab:computational_cost} reports the computational cost of the above four models. Compared to cscSE-UNet, our models have a reduced number of parameters and a reduced amount of training time but slightly increased inference time. Our maximum training time was around six hours, and the inference time is within seconds for a subject. We tried experiments with a 3D U-Net and a convolutional LSTM; however, we met memory difficulties on our machine, which motivates the MDA-Net.

\section{Conclusion and Discussion}
This paper investigated the 3D image segmentation problem and proposed an efficient solution using a multi-dimensional attention network. We tested the network on image volumes; however, it could be extended to handle the spatiotemporal data like videos or longitudinal images. We will apply our model to other segmentation tasks with different image types in the future. For multi-modality image scans, we could explore image-wise attention, that is, measuring the contributions of each modality to the segmentation task.

\section*{\fontsize{12}{15}\selectfont Compliance with Ethical Standards} This is a numerical simulation study for which no ethical approval was required. 

\section*{\fontsize{12}{15}\selectfont Acknowledgements} No funding was received for conducting this study. The authors have no relevant financial or non-financial interests to disclose.

%the use of a modified Squeeze and Excitation concept along the volume dimension, and to the best of our knowledge this is the first application of this method in this way and the results were better than simply using a 2D slice. Our modified SE block performs an efficient Depthwise Operation by first flattening the 3D volume and then applying one depthwise filter to all the channels stacked side by side.
%For each slice in the 3-D volume, the difference and order operators rearrange the remainder of the 3-D volume in order of importance with respect to the current slice. However this is static and the Squeeze and Excitation blocks enhance and provide a dynamic attention by further assigning a weight to the input slices. The Conv 1x1 in turn compresses this information into a single slice and provides it as a input which is then paired with the current slice and sent to the encoder-decoder section. In future we hope to extend this concept along the time dimension, and apply it on 3D+T data to replace LSTM cost overheads. 

% \section{References}
\bibliographystyle{IEEEbib}
\bibliography{isbi_ref}

\end{document}